\documentclass{article}

\PassOptionsToPackage{numbers,sort&compress,comma}{natbib}
\usepackage{arxiv}

\usepackage[utf8]{inputenc}
\usepackage[T1]{fontenc}
\usepackage{hyperref}
\usepackage{url}
\usepackage{booktabs}
\usepackage{amsfonts}
\usepackage{amssymb}
\usepackage{nicefrac}
\usepackage{microtype}
\usepackage{graphicx}
\usepackage{doi}
\usepackage{amsmath}

\title{
Momentum-Anchored Multi-Scale Fusion Model for Long-Tailed Chest X-Ray Classification
\thanks{An earlier version of this work was accepted at the FPT International Conference on Emerging Trends in Computing (FETC).}
}

\author{
\href{mailto:duyhkse184883@fpt.edu.vn}{Hoang Khuong Duy} \\
Department of Artificial Intelligence\\
FPT University\\
Ho Chi Minh City, Vietnam\\
\texttt{duyhkse184883@fpt.edu.vn}
\And
\href{mailto:duynhse183995@fpt.edu.vn}{Nguyen Huu Duy} \\
Department of Artificial Intelligence\\
FPT University\\
Ho Chi Minh City, Vietnam\\
\texttt{duynhse183995@fpt.edu.vn}
\And
\href{mailto:nguhcv@fe.edu.vn}{Huynh Cong Viet Ngu\thanks{Correspondence can be addressed to Huynh Cong Viet Ngu at \texttt{nguhcv@fe.edu.vn}.}} \\
Department of Computing Fundamental\\
FPT University\\
Ho Chi Minh City, Vietnam\\
\texttt{nguhcv@fe.edu.vn}
}

\renewcommand{\shorttitle}{\textit{arXiv} Template}

\hypersetup{
pdftitle={A template for the arxiv style},
pdfsubject={q-bio.NC, q-bio.QM},
pdfauthor={David S.~Hippocampus, Elias D.~Striatum},
pdfkeywords={First keyword, Second keyword, More},
}

\begin{document}
\maketitle

\begin{abstract}
Chest X-ray classification suffers from severe class imbalance where gradient updates bias toward majority classes, causing feature drift and poor performance on rare but critical pathologies. We propose a Momentum-Anchored Multi-Scale Fusion Network that uses exponential moving averages (EMA) as a temporal anchoring mechanism to stabilize feature representations under long-tailed distributions. Our approach applies selective momentum updates to the final expansion block of an EfficientNet backbone, creating a slowly-evolving reference branch that resists gradient-induced drift while preserving discriminative patterns for minority classes. Combined with multi-scale spatial fusion (1×1, 3×3, 5×5 convolutions), this anchoring strategy maintains representational stability throughout training. On ChestX-ray14, our method achieves 0.8682 average AUC, outperforming state-of-the-art approaches and showing particular improvements on rare pathologies like Hernia (0.9470) and Pneumonia (0.8165). The results demonstrate that momentum anchoring effectively counters feature instability in long-tailed medical image classification.
\end{abstract}

\noindent\textbf{Keywords}: Memory-augmented neural networks, chest X-ray, class imbalance, deep learning, medical imaging

\section{Introduction}

\begin{figure}[h]
    \centering
    \includegraphics[width=0.5\textwidth]{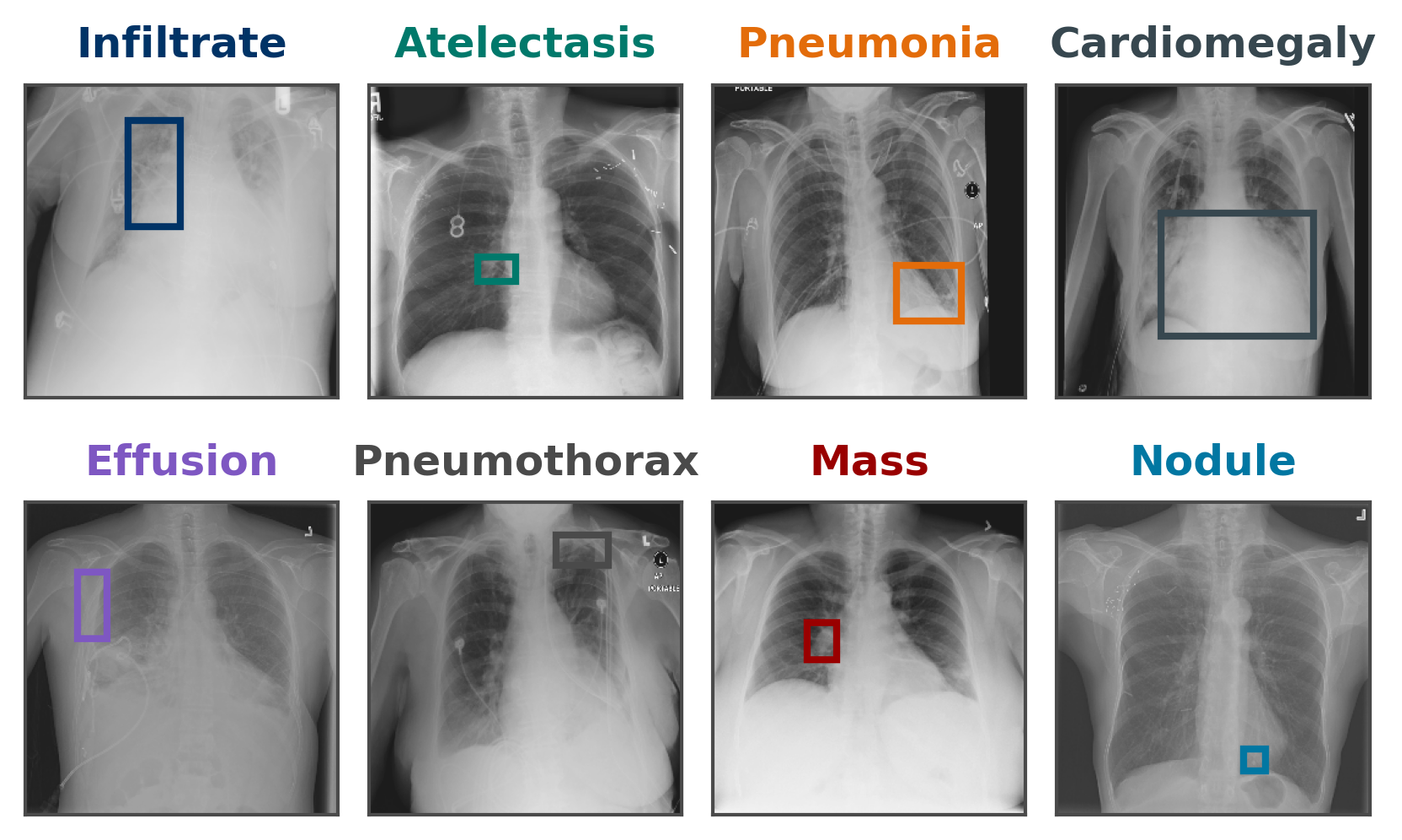}
    \caption{Disease Localization in Chest X-Ray Images.}
    \label{fig:localization}
\end{figure}

Automated chest X-ray (CXR) interpretation has emerged as a critical application of deep learning in medical imaging, with the potential to assist radiologists in detecting multiple co-occurring thoracic pathologies. While convolutional neural networks have demonstrated impressive performance on large-scale CXR datasets~\cite{anwar2018medical, wang2017chestx, rajpurkar2017chexnet}, real-world medical datasets present a dual challenge: severe long-tailed distributions where common conditions vastly outnumber rare but clinically critical pathologies, combined with extreme spatial variability in disease manifestations (Fig.~\ref{fig:localization}). Pathologies exhibit significant variation in size, shape, location, and radiographic appearance across patients—from subtle nodules requiring fine-grained texture analysis to large consolidations spanning multiple lung regions. This combination of statistical imbalance and morphological diversity creates a cascade of learning challenges that traditional approaches struggle to address effectively.

The core challenge lies in the compounded effect of class imbalance and disease heterogeneity on gradient-based optimization. In long-tailed medical datasets, frequent pathologies such as Cardiomegaly or Effusion not only generate substantially more gradient updates than rare conditions like Hernia or Pneumothorax~\cite{holste2024towards, johnson2019survey}, but also exhibit more consistent spatial patterns that dominate the learning process. Meanwhile, rare pathologies often present with high morphological variability—appearing as diffuse patterns, focal lesions, or subtle textural changes that require different spatial receptive fields for effective detection. This dual challenge of statistical and morphological imbalance causes systematic feature drift, where learned representations progressively shift toward spatially consistent patterns of common diseases while simultaneously losing the fine-grained discriminative features essential for detecting morphologically diverse rare pathologies. Consequently, models achieve high overall accuracy by correctly classifying frequent cases but fail catastrophically on minority classes that often represent the most urgent clinical scenarios.

Existing approaches to class imbalance—including spatial data augmentation~\cite{chlap2021review}, and loss reweighting~\cite{jin2023deep, chamseddine2022handling}—primarily address the statistical symptom rather than the underlying cause of representational instability under morphological diversity. While these methods can improve minority class recall, they do not fundamentally solve the gradient-induced feature drift that occurs throughout the training process, nor do they account for the spatial heterogeneity within each disease category. Attention mechanisms~\cite{tian2023unsupervised,wang2021triple,yang2024feature} and multi-scale feature fusion~\cite{lin2024multibranch,ashraf2023synthensemble}, though effective for capturing spatial localization across different disease presentations, similarly fail to maintain temporal consistency in learned representations when faced with the combined challenge of severely imbalanced training signals and highly variable disease morphologies. The spatial variability of rare pathologies—ranging from tiny nodules to extensive infiltrations—requires stable multi-scale feature learning that existing approaches cannot guarantee under gradient drift.

Recent advances in self-supervised learning have demonstrated the power of momentum-based encoders, particularly exponential moving averages (EMA)\cite{izmailov2018averaging}, for creating stable and robust feature representations. Originally developed for contrastive learning frameworks like MoCo\cite{he2020momentum} or BYOL\cite{grill2020bootstrap}, momentum mechanisms maintain slowly-evolving target networks that provide consistent training signals. However, these approaches have not been systematically adapted to address the specific challenge of feature stability in long-tailed medical classification.

In this work, we introduce a novel momentum anchoring strategy specifically designed to counter gradient-induced feature drift while simultaneously preserving multi-scale discriminative patterns for morphologically diverse pathologies in imbalanced medical datasets. Our key insight is that by selectively applying momentum updates to high-capacity layers, we can create a temporal anchor that preserves both fine-grained textural features for subtle lesions and broader contextual patterns for extensive pathologies, while allowing rapid adaptation for common conditions. This approach differs fundamentally from uniform momentum schemes by targeting the layers most susceptible to feature drift and integrating multi-scale spatial processing to handle disease morphological diversity.

Our main contributions are:
\begin{itemize}
    \item \textbf{Momentum Anchoring Mechanism:} We propose a selective momentum-based update strategy applied to the final expansion block of deep networks, creating a temporally stable reference branch that resists gradient-induced drift caused by class imbalance.
    \item \textbf{Multi-Scale Spatial Integration:} We develop a hierarchical fusion module that combines momentum-anchored features with multi-scale spatial representations (1×1, 3×3, 5×5 convolutions), addressing both temporal stability and spatial variability challenges simultaneously.
    \item \textbf{Theoretical Analysis:} We provide mathematical insights into how momentum anchoring implicitly smooths historical gradients, demonstrating its effectiveness in maintaining feature stability under long-tailed distributions.
    \item \textbf{Empirical Validation:} Comprehensive experiments on ChestX-ray14 demonstrate significant improvements in rare pathology detection, with our method achieving 0.8682 average AUC and outperforming state-of-the-art approaches in 13 out of 14 disease categories.
\end{itemize}

By combining multi-scale fusion with momentum-guided memory updates, our model addresses class imbalance and improves rare disease detection in multi-label chest X-ray classification.

The remainder of this paper is organized as follows: Section 2 reviews related work. Section 3 introduces the proposed method. Section 4 details the experimental setup. Section 5 presents the results and analysis. Finally, Section 6 concludes the paper and discusses future directions.

 \section{Related Work}

\subsection{Chest X-ray Classification and Deep Learning}

Chest X-ray classification has advanced significantly with deep learning, particularly Convolutional Neural Networks (CNNs). Early models such as CheXNet~\cite{rajpurkar2017chexnet} achieved radiologist-level performance in pneumonia detection, catalyzing multi-label thoracic disease classification using large-scale datasets like ChestX-ray14~\cite{wang2017chestx}. Subsequent efforts have leveraged architectures including DenseNet~\cite{huang2017densely}, EfficientNet~\cite{tan2019efficientnet}, and Vision Transformers~\cite{dosovitskiy2020image} to enhance accuracy and robustness. Nonetheless, challenges persist due to class imbalance and the subtle nature of certain pathologies~\cite{anwar2018medical}, motivating continued research in advanced feature extraction and representation learning.

\subsection{Localization Challenges in Chest X-ray Diagnosis}

Accurate localization of pathological regions in chest X-rays is critical for both clinical interpretability and enhancing multilabel classification performance. However, localization remains a challenging problem due to the high variability in lesion appearance, size, and location across patients \cite{wang2017chestx}. Weakly supervised learning approaches, which leverage image-level labels without requiring bounding box annotations, have been widely explored \cite{wang2017chestx, tian2023unsupervised}. Attention mechanisms \cite{wang2021triple, guan2018diagnose} and multi-resolution feature fusion \cite{yang2024feature, li2024review} have shown promise in improving localization accuracy. Nevertheless, existing methods often exhibit bias toward common disease patterns and fail to generalize well to rare or atypical presentations, limiting their clinical utility \cite{mao2022imagegcn, tian2023unsupervised}. This persistent limitation underscores the need for more robust and adaptive localization frameworks.

\subsection{Momentum-based Approaches}

Momentum-based encoders leverage exponential moving averages (EMA)~\cite{izmailov2018averaging} to update a target network from an online encoder, providing a stable yet adaptive signal for representation learning. Originally introduced in MoCo~\cite{he2020momentum}, this paradigm maintains a large dictionary of momentum-updated embeddings for contrastive learning. MoCo-CXR~\cite{sowrirajan2021moco} adapted this framework to chest X-rays, showing that MoCo-pretrained ResNet-18 encoders outperform ImageNet and randomly initialized baselines, particularly in low-label settings on CheXpert and external tuberculosis datasets.

BYOL~\cite{grill2020bootstrap} further advances this approach by eliminating negative pairs, training an online encoder to predict the projections of a momentum-updated target. While BYOL achieves strong results in natural image and audio domains, it has not been explored for multi-label chest X-ray classification.

Existing self-supervised medical imaging methods—including MoCo-CXR, MedAug~\cite{vu2021medaug}, and ConVIRT~\cite{zhang2022contrastive}—typically apply a uniform momentum rate across all layers. This uniformity may oversmooth fine-grained features crucial for rare pathologies, limiting sensitivity to minority classes. To address this, we propose a block-wise momentum update strategy that assigns higher momentum to shallow layers for anchoring low-level features, while allowing deeper layers to adapt more rapidly. This selective design improves representation learning for long-tailed disease distributions in chest X-ray classification.

\section{Proposed Model for Long-Tailed Chest
X-Ray Classification}
\subsection{Overall Architecture}
\begin{figure}[h]
\centering
\includegraphics[width=1.0\textwidth]{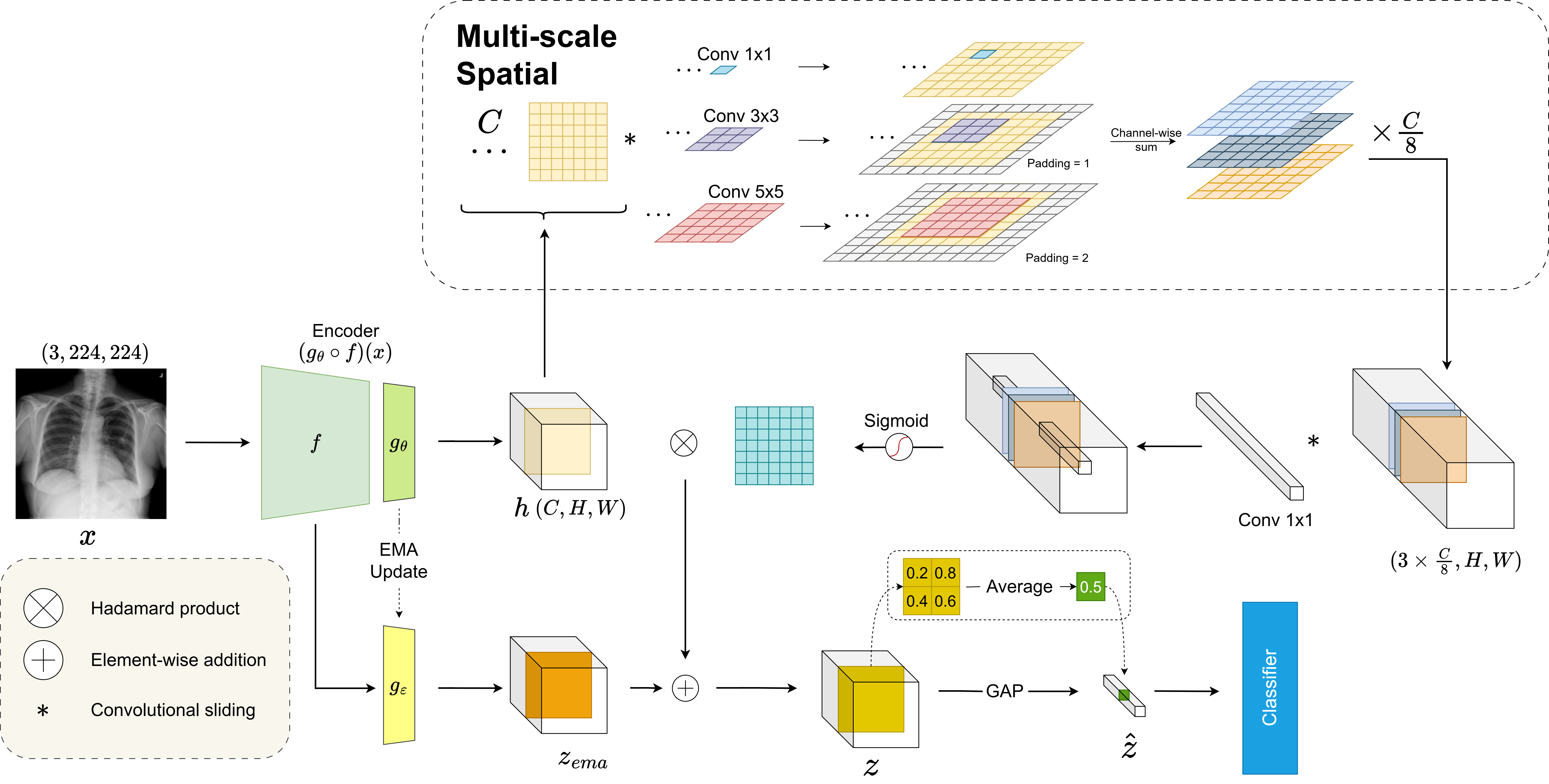}
    \caption{\textbf{Overview of the dual‐path feature extraction framework}. The input chest X-ray image $x$ is first processed by an EfficientNetV2-S backbone (encoder $g_\theta\circ f$) to produce high-level feature maps $h$. From the final block of this encoder, a momentum branch $g_{\theta}$ is forked and its weights are updated via EMA, yielding stabilized embeddings $z_{\mathrm{ema}}$. Simultaneously, the primary encoder output $h$ is fed into a hierarchical fusion module that applies parallel $1\times1$, $3\times3$, and $5\times5$ convolutions to generate multi-scale attention features $z_{\mathrm{att}}$. The momentum‐stabilized features and fused multi-scale features are then combined to form the final representation $z$, which undergoes global average pooling (GAP) before classification.}
    \label{fig:overall_architecture}
\end{figure}

To address the substantial spatial variability of disease manifestations in chest X-rays, we introduce a dual-path feature extraction framework (Fig.~\ref{fig:overall_architecture}). An EfficientNetV2-S backbone generates high-level feature maps, from which a momentum-updated branch is forked at the final expansion block. This auxiliary branch maintains weights via an exponential moving average (EMA) of the main encoder, producing a slowly evolving representation stream that anchors learning and enhances stability for subtle, spatially diverse lesions.

Concurrently, the main encoder output is passed to a Momentum-Anchored Multi-Scale Fusion module, comprising parallel $1{\times}1$, $3{\times}3$, and $5{\times}5$ convolutions to extract features at multiple spatial resolutions. Fusing the momentum-stabilized embeddings with the multi-scale features enhances both localization robustness and sensitivity to fine-grained patterns, improving classification of pathologies with high positional variance across patients.

\subsection{Feature Extraction}

Let the input chest X‐ray be \(x \in \mathbb{R}^{3\times H\times W}.\)
Our EfficientNetV2‐S backbone continuously transforms this input via an initial feature extractor \(f\) and then an expansion block \(g_{\theta}\), yielding
\[
x^{(1)} = f(x)\;\in\;\mathbb{R}^{D\times H'\times W'}, 
\qquad
h = g_{\theta}(x^{(1)})\;\in\;\mathbb{R}^{C\times H''\times W''}
\tag{1}
\]
where \(f\) is the feature extractor (outputting \(C=256\) channels) and \(g_{\theta}\) is the final expansion block parameterized by \(\theta\) (outputting \(D=1280\) channels).

\subsection{Momentum Path}

To mitigate instability in high‐dimensional feature learning—especially for rare or underrepresented pathologies—we maintain a secondary “momentum” branch that mirrors the final expansion block of the EfficientNetV2‐S backbone.  Concretely, let \(x^{(1)}\in\mathbb{R}^{C\times H'\times W'}\) be the feature map immediately preceding the expansion block \(g_{\theta}\).  We define a momentum‐updated block \(g_{\epsilon}\) with parameters \(\epsilon\), which processes the same input to yield
\[
z_{\mathrm{ema}} 
\;=\; g_{\epsilon}\bigl(x^{(1)}\bigr)
\;\in\;\mathbb{R}^{C\times H''\times W''}.
\tag{2}
\]
Here, \(\epsilon\) is initialized to the encoder’s weights, \(\epsilon_{0} = \theta_{0}\), and updated after each mini‐batch via an exponential moving average:
\[
\epsilon_{t} 
\;=\; m\,\epsilon_{t-1} \;+\; (1-m)\,\theta_{t-1},
\quad m\in[0,1),
\tag{3}
\]
where \(m\) is the momentum coefficient.  This slow‐moving target branch “anchors” the representation space, reducing the impact of noisy gradient updates and preserving information about infrequent lesion patterns.

Moreover, one can unroll the EMA update to reveal its implicit smoothing over past gradients.  By substituting \(\theta_{t} = \theta_{t-1} - \eta\,\nabla_{\theta_{t-1}}\mathcal{L}\), we obtain the approximation
\[
\epsilon_{t}
\;\approx\;
\theta_{t}
\;+\;
\sum_{i=1}^{t-1}
m^{\,i}\,\eta\,\nabla_{\theta_{t-i}}\mathcal{L}
\;+\;
m^{\,t}\,(\epsilon_{1}-\theta_{1}),
\tag{4}
\]
where \(\eta\) is the learning rate and \(\nabla_{\theta_{t-i}}\mathcal{L}\) the gradient at step \(t-i\).  Equation (5) elucidates how the momentum branch integrates exponentially decayed historical gradients, promoting smoother parameter trajectories and more stable batch‐normalization statistics.

By targeting only the expansion block, this dual‐path mechanism imposes minimal additional compute, while effectively “slow‐locks” high‐capacity features against erratic updates—thereby enhancing sensitivity to rare, variably localized pathologies without impeding rapid adaptation in earlier layers.  
\subsubsection{Analysis on Slowing effects of Momentum Branch}
\subsection{Momentum-Anchored Multi-Scale Fusion}

To directly address the challenge of extreme spatial variability in lesion appearance, we introduce a Momentum-Anchored Multi-Scale Fusion that integrates multi–scale spatial cues with temporally stabilized embeddings. Given
\[
 h \in \mathbb{R}^{C\times H''\times W''},
 z_{\mathrm{ema}} \in \mathbb{R}^{C\times H''\times W''},
\]
we first extract scale–diverse responses via parallel convolutions and then sum and project them:

\begin{equation}
z_{\mathrm{att}}
= W'_{1\times1} \ast
(  W_{1\times1}\ast h  + W_{3\times3}\ast h  + W_{5\times5}\ast h ),
\tag{1}
\end{equation}
where each branch reduces channels to $\frac{C}{8}$. Concretely, the three parallel convolutions each output a tensor of shape $(\frac{C}{4},H'',W'')$. We then perform a channel-wise summation across these three responses for each of the \(\frac{C}{8}\) groups, collapsing them into single-channel feature maps. Repeating this process \(\frac{C}{8}\) times yields an intermediate tensor of shape $(3\times\frac{C}{8},H''×W'')$. Finally, this tensor is passed through the projection convolution $W'_{1×1}$ to produce $z_{att}$ with the desired channel dimension \(\frac{C}{8}\) \cite{szegedy2015going}. This design enforces direct interaction across scales, ensuring fine textures and broader context are jointly modeled.

Next, we merge these fused features with the slow–evolving momentum stream:

\begin{equation}
z = z_{\mathrm{att}} + z_{\mathrm{ema}},
 m \in \mathbb{R}^{C\times H''\times W''},
\tag{2}
\end{equation}

thereby anchoring multi–scale activations against noisy positional shifts. Critically, this addition before pooling guarantees that spatially varying cues and temporally consistent patterns interact at every location.

Finally, a single global average pooling over \(m\) produces the compact representation:

\begin{equation}
\hat{z}_{c} = \frac{1}{H''W''}
  \sum_{i=1}^{H''}\sum_{j=1}^{W''} z_c(i,j),
 c = 1,\dots,C.
\tag{3}
\end{equation}

By integrating these two streams prior to pooling, our Momentum-Anchored Multi-Scale Fusion directly mitigates localization noise while preserving the discriminative detail required to detect rare or subtly manifest pathologies.

The output of the global pooling, $\hat z$, serves as the input to the classification head. This head is a multi-layer perceptron $g$ that applies batch normalization to the pooled feature vector, followed by a linear transformation projecting it to 512 dimensions. A ReLU activation and a dropout with rate 0.3 are then applied for non-linearity and regularization, and a final linear layer outputs the raw logits for the $K$ target classes, representing the model’s unnormalized confidence scores.

\section{Experimental Setup and Training Process}

\textbf{Dataset and Preprocessing.} We evaluate our method on the ChestX-ray14 dataset~\cite{wang2017chestx}, comprising 112{,}120 frontal-view radiographs labeled with 14 thoracic diseases. Images are resized to $224 \times 224$ and normalized using ImageNet statistics. Following~\cite{wang2017chestx}, we split the dataset into 70\% training, 10\% validation, and 20\% test sets using three different random seeds.

\textbf{Training Protocol.} We adopt the two-stage training procedure from SynthEnsemble~\cite{ashraf2023synthensemble}, modifying only the loss functions and batch sizes. Stage 1 uses Binary Cross-Entropy loss on images with at least one positive label (batch size 64). Stage 2 trains on the full dataset using Asymmetric Loss with batch size increased to 128.

\textbf{Evaluation.} We report the area under the ROC curve (AUC) per class on the validation set as the primary performance metric.

\section{Results}
This section presents the experimental results of the proposed model on the ChestX-ray14 dataset, including a comparison of its performance with state-of-the-art methods and an ablation study to evaluate each architectural component.
\subsection{Comparison with State-of-the-Art Methods}
Table \ref{tab:comparison} provides a detailed comparison of the Area Under the Curve (AUC) scores of the momentum-anchored multi-scale fusion network model against leading methods. The proposed model achieved an average AUC of 0.8682, setting a new performance benchmark and surpassing the previously strongest method, SynthEnsemble (0.8543), by a significant margin of 1.39 percentage points.

When analyzing each pathology in detail, the model showed overall superiority, ranking first in 13 out of 14 pathologies. Notable improvements were recorded in challenging pathologies such as Hernia (0.9470), Pneumothorax (0.9192), and Edema (0.9301). The only pathology where the model did not rank first was Infiltration, where it achieved 0.7343, placing it behind SynthEnsemble (0.74102). The values in parentheses for our method represent the confidence interval (minimum, maximum) across different runs, indicating the stability of the results.

\begin{table}[h]
\centering
\label{tab:comparison}
\resizebox{\textwidth}{!}{
\begin{tabular}{|l|c|c|c|c|c|c|c|c|}
\hline
\textbf{Pathology} & \textbf{Wang.\cite{wang2017chestx}} & \textbf{CheX\cite{rajpurkar2017chexnet}} & \textbf{Synth\cite{ashraf2023synthensemble}} & \textbf{A$^3$ Net\cite{wang2021triple}} & \textbf{IGCN\cite{mao2022imagegcn}} & \textbf{Taslimi.\cite{taslimi2022swinchex}} & \textbf{Manzari.\cite{manzari2023medvit}} & \textbf{Ours} \\
\hline
Atelectasis       & 0.700 & 0.8094 & \underline{0.83390} & 0.779 & 0.802 & 0.781 & - & \textbf{0.8355} $\pm$0.0072 \\
Cardiomegaly      & 0.810 & \underline{0.9248} & 0.91954 & 0.895 & 0.894 & 0.875 & - & \textbf{0.9346} $\pm$0.0050 \\
Effusion          & 0.759 & 0.8638 & \underline{0.88977} & 0.836 & 0.874 & 0.824 & - & \textbf{0.9021} $\pm$0.0014 \\
Infiltration      & 0.661 & \underline{0.7345} & \textbf{0.74102} & 0.710 & 0.709 & 0.701 & - & 0.7343 $\pm$0.0052 \\
Mass              & 0.693 & 0.8676 & \underline{0.87315} & 0.834 & 0.843 & 0.822 & - & \textbf{0.8881} $\pm$0.0029 \\
Nodule            & 0.669 & 0.7802 & \underline{0.80611} & 0.777 & 0.768 & 0.780 & - & \textbf{0.8165} $\pm$0.0017 \\
Pneumonia         & 0.658 & 0.7680 & \underline{0.77648} & 0.737 & 0.715 & 0.713 & - & \textbf{0.7903} $\pm$0.0150 \\
Pneumothorax      & 0.799 & 0.8887 & \underline{0.90164} & 0.878 & 0.900 & 0.871 & - & \textbf{0.9192} $\pm$0.0051 \\
Consolidation     & 0.703 & 0.7901 & \underline{0.81575} & 0.759 & 0.796 & 0.748 & - & \textbf{0.8290} $\pm$0.0125 \\
Edema             & 0.805 & 0.8878 & \underline{0.91034} & 0.855 & 0.883 & 0.848 & - & \textbf{0.9301} $\pm$0.0094 \\
Emphysema         & 0.833 & 0.9371 & 0.92946 & \underline{0.933} & 0.915 & 0.914 & - & \textbf{0.9445} $\pm$0.0076 \\
Fibrosis          & 0.786 & 0.8047 & \underline{0.83347} & 0.838 & 0.825 & 0.826 & - & \textbf{0.8539} $\pm$0.0026 \\
Pleural Thickening & 0.684 & 0.8062 & \underline{0.81270} & 0.791 & 0.791 & 0.778 & - & \textbf{0.8300} $\pm$0.0136 \\
Hernia            & 0.872 & 0.9164 & 0.91723 & 0.938 & \underline{0.943} & 0.855 & - & \textbf{0.9470} $\pm$0.0319 \\
\hline
\textbf{Mean AUC} & 0.745 & 0.841 & \underline{0.85433} & 0.826 & 0.832 & 0.810 & 0.805 & \textbf{0.8682} $\pm$0.0020 \\
\hline
\end{tabular}
}
{\footnotesize {Values for our method show mean (min, max) across multiple runs. \textbf{Bold} indicates the highest value; \underline{underlined} indicates the second-highest value in each row.}}
\caption{Comparison of AUC scores with state-of-the-art methods on the ChestX-ray14 dataset.}
\label{tab:comparison}
\end{table}

\subsection{Ablation Study}

To determine the contribution of key components, an ablation study was conducted, and the results are presented in Table 2. The baseline model, without the multi-scale fusion module and momentum mechanism, achieved an average AUC of 0.8304.


\begin{table}[h]
\label{tab:ablation}
\resizebox{\textwidth}{!}{
\begin{tabular}{|c|c|c|c|c|}
\hline
\textbf{Multi-scale fusion Module} & \textbf{Momentum (m=0.999)} & \textbf{Mean AUC}       & {\textbf{GFLOPS}} & \textbf{Total Parameters} \\
\hline
                       $\times$            &           $\times$        & 0.8304 $\pm$0.0119 & 2.901G          & 20.843M                   \\
                       
                       \hline
                     $\checkmark$              &        $\times$           & 0.8613 $\pm$0.0056 & 3.253G          & 28.012M \\
                     
                     \hline
                       $\times$            & $\checkmark$             & 0.8670 $\pm$0.0038 & 2.918G                              & 21.173M                   \\
                       \hline
                     $\checkmark$              & $\checkmark$             & \textbf{0.8682 $\pm$0.0020} & 3.629G                              & 28.342M                   \\
                     \hline
\end{tabular}}
\caption{Ablation Study of Multi-scale Fusion and Momentum Parameters.}
\end{table}

Upon integrating the Multi-scale Fusion Module, performance increased sharply to 0.8613 (without momentum), demonstrating the core role of synthesizing information from features at multiple scales. Subsequently, the addition of the Momentum Anchoring mechanism further improved the results. With a momentum coefficient of m=0.999, the model achieved a final performance of 0.8682. This shows the sequential and significant contribution of both proposed components, with the multi-scale module providing the main performance leap and the momentum mechanism helping to fine-tune for optimal performance.

\subsection{Computational Efficiency}

In addition to diagnostic performance, the computational cost of the model was also evaluated. As shown in Table 2, the complete model requires 3.629 GFLOPS and contains 28.342 million parameters. Although these metrics are higher than those of the baseline model, the increase is reasonable given the significant improvement of nearly 3.8\% AUC points. This indicates that the model achieves an effective balance between accuracy and practical deployment feasibility.

\section{Conclusion}

This paper presents a momentum-anchored multi-scale fusion network to address class imbalance in chest X-ray classification. Our model selectively applies momentum updates, preserving representations of rare pathologies without significant computational cost. Its dual-path architecture integrates stable temporal features with multi-scale spatial cues, enabling robust detection across diverse lesion types.

While our approach achieved a mean AUC of 0.8682 on ChestX-ray14, outperforming previous methods in 13 of 14 categories, a notable limitation is its second-place performance in detecting "Infiltration." This is likely due to the diffuse, textural nature of the pathology, which poses a challenge for standard convolutional methods.

Future work could address this by integrating specialized attention mechanisms to better focus on subtle patterns or by exploring ensemble strategies that combine our model with one specializing in textural analysis. Additionally, leveraging external datasets for pre-training could further enhance the model's foundational feature extraction capabilities.



\bibliographystyle{unsrt}
\bibliography{references}
\end{document}